\documentclass{article}

\usepackage{amsmath}
 \usepackage[dblblindworkshop, final]{neurips_2025}
\workshoptitle{2nd Workshop on Multi-modal Foundation Models and Large Language Models for Life Sciences}

\usepackage[utf8]{inputenc} 
\usepackage[T1]{fontenc}    
\usepackage{hyperref}       
\usepackage{url}            
\usepackage{booktabs}       
\usepackage{amsfonts}       
\usepackage{nicefrac}       
\usepackage{subcaption}
\usepackage{microtype}      
\usepackage{xcolor}         
\usepackage{graphicx}
\usepackage{float}                 
\usepackage{placeins}

\FloatBarrier

\title{Thin Bridges for Drug Text Alignment: Lightweight Contrastive Learning for Target Specific Drug Retrieval}

\author{%
  Mallikarjuna Tupakula \thanks{Use footnote for providing further information
    about author (webpage, alternative address)---\emph{not} for acknowledging
    funding agencies.} \\
  Rochester Institute of Technology\\
  Rochester, NY 14623 \\
  \texttt{mt3998@rit.edu} \\\texttt{tmallikarjuna111@gmail.com} \\
}

\begin{document}

\maketitle

\begin{abstract}
Multimodal foundation models hold promise for drug discovery and biomedical applications, but most existing approaches rely on heavy pretraining or large scale multimodal corpora. We investigate whether \textit{thin contrastive bridges}, lightweight projection heads over frozen unimodal encoders can align chemical and textual representations without training a full multimodal model. Using paired mechanisms from ChEMBL, we align \textit{ECFP4 molecular fingerprints} with biomedical sentence embeddings through dual linear projections trained with a contrastive objective. To better handle drugs sharing the same therapeutic target, we incorporate hard negative weighting and a margin loss. Evaluation under scaffold based splits, which require generalization across disjoint chemical cores, demonstrates that our approach achieves non-trivial cross modal alignment and substantially improves within target discrimination compared to frozen baselines. These results suggest that thin bridges offer a compute efficient alternative to large scale multimodal pretraining, enabling scaffold aware drug text alignment and target specific retrieval in precision medicine.
\end{abstract}

\section{Introduction}
Multimodal foundation models have opened promising directions for biomedical research, particularly in drug discovery and precision medicine. These approaches seek to unify heterogeneous representations such as chemical structures, protein targets, and biomedical text under a shared embedding space. Recent advances in contrastive multimodal learning have demonstrated that cross-modal alignment can support drug target prediction, phenotype based screening, and chemical structure elucidation \cite{xu2023asymmetric}, \cite{wang2025advancing}, \cite{rao2025multi, rocabert2025multi}. For instance, CLOOME showed that contrastive learning can unlock bioimaging databases for queries with chemical structures \cite{sanchez2023cloome}, while MolCLR and related efforts emphasized molecular contrastive learning as an efficient path toward transferable molecular representations \cite{wang2022molecular, pinheiro2022smiclr}. Beyond supervised pipelines, lightweight frameworks such as OneEncoder \cite{faye2024oneencoder} and cross-modal efficiency strategies \cite{faye2024oneencoder} suggest that thin projection heads over frozen encoders can serve as compute efficient bridges across modalities.

A growing line of work extends these retrieval based strategies toward drug design and discovery applications. Efforts such as multimodal protein ligand contrastive learning \cite{wang2024enhancing}, retrieval augmented biomedical learning \cite{he2025retrieval, gargari2025enhancing}, and unified multimodal pipelines \cite{luo2024toward, dang2025multimodal} highlight the importance of efficiently linking chemical fingerprints with textual or biological context for downstream drug discovery tasks. Importantly, such methods emphasize that retrieval accuracy is not merely an evaluation metric but a foundation for enabling generation whether for novel molecule design or phenotype conditioned predictions \cite{huang2024unified}.

Parallel insights arise from \textbf{neuroscience} and \textbf{brain decoding}, where retrieval aligned models have been shown to enhance downstream generative reconstruction. For example, THINGS-data \cite{hebart2023things} provides a multimodal benchmark of fMRI, MEG, and behavioral similarity judgments, enabling representational alignment across modalities. Similarly, \cite{defossez2023decoding} demonstrated that contrastive alignment of MEG/EEG signals with pretrained speech embeddings supports zero-shot decoding of perceived speech segments with high accuracy. In vision, \cite{benchetrit2023brain} and related work on fMRI-to-image decoding showed that improved retrieval alignment between brain signals and latent image embeddings leads to higher fidelity image reconstructions when paired with diffusion based generative models. These findings reinforce the idea that retrieval accuracy is tightly coupled with generative quality, motivating analogous strategies in biomedical domains.

Taken together, these advances suggest that compute efficient cross modal retrieval architectures can serve as scalable building blocks for more ambitious generative models. In this work, we investigate whether thin projection bridges lightweight contrastive heads over frozen unimodal encoders can align chemical fingerprints with biomedical mechanism text, while disambiguating drugs with shared targets through hard-negative sampling and margin based losses. By evaluating under scaffold splits on ChEMBL\cite{gaulton2012chembl}, we demonstrate non-trivial generalization and within target retrieval gains, supporting the broader view that such \textit{thin bridges} can form the foundation for downstream generative pipelines in precision drug discovery.

\paragraph{Contributions.}
This paper makes the following contributions:
\begin{itemize}
    \item We introduce \textbf{thin contrastive bridges} that align ECFP4 molecular fingerprints with biomedical mechanism text through lightweight dual projection heads, avoiding large-scale multimodal pretraining.
    \item We incorporate \textbf{hard-negative weighting and a margin based loss} to better disambiguate drugs that act on the same therapeutic target, addressing within target retrieval challenges.
    \item We evaluate on ChEMBL with a rigorous \textbf{scaffold-based split}, demonstrating non-trivial cross modal generalization and improved within target retrieval compared to frozen encoder baselines.
    \item We show that such thin bridges are \textbf{compute-efficient} (single GPU, short training time) and can serve as scalable foundations for downstream generative drug discovery pipelines.
\end{itemize}

\section{Dataset}

We constructed our dataset from \textbf{ChEMBL v28}, a curated database of bioactive molecules with drug like properties. 
To ensure clinical relevance, we extracted all \emph{approved drugs} by filtering entries with \texttt{max\_phase = 4}. 
From the \texttt{mechanism} table, we collected associations between drugs and their therapeutic targets, retaining only 
records with explicit molecular mechanisms. 

To enrich these entries, we combined three information sources:
\begin{itemize}
    \item \textbf{Molecule data}: Canonical SMILES strings retrieved from the ChEMBL molecule endpoint, corresponding to 
    2,970 unique molecules after deduplication.
    \item \textbf{Mechanism data}: Textual descriptions of drug mechanisms and annotated \emph{action types} 
    (e.g., inhibitor, agonist).
    \item \textbf{Target data}: Standardized ChEMBL target identifiers, mapped to preferred names and biological target types 
    via the target endpoint.
\end{itemize}

To reduce redundancy, we removed duplicate drug target pairs and dropped incomplete rows lacking either SMILES or  mechanism text. After filtering, the resulting dataset comprised \textbf{3,030 high quality drug–target pairs} with both 
chemical and textual representations. Each entry contains:
\begin{itemize}
    \item Molecule ChEMBL ID
    \item Canonical SMILES
    \item Mechanism of action (free-text biomedical description)
    \item Target ChEMBL ID and target name
    \item Action type
    \item Maximum clinical phase
\end{itemize}

This dataset provides a paired multimodal resource aligning molecular fingerprints with biomedical text.  The SMILES strings offer a cheminformatic view of drug structure, while the mechanism sentences encapsulate human curated  biomedical knowledge, often specifying target level interactions. This dual representation makes the dataset well suited  for \textbf{contrastive drug text alignment} tasks, where the objective is to learn lightweight bridges between unimodal  embeddings. 

By grounding the dataset in ChEMBL, a widely adopted benchmark in drug discovery, we ensure both interpretability and  extensibility. The final corpus balances \emph{domain richness} (multiple therapeutic areas, diverse targets) with  \emph{computational tractability} (single GPU scale), enabling efficient exploration of multimodal alignment methods for drug retrieval and, ultimately, generative drug design.

\section{Methodology}

\subsection{Dataset Construction}
We use drug target pairs from ChEMBL \cite{gaulton2012chembl}, restricted to 
approved drugs (\texttt{max\_phase = 4}). Each entry includes canonical SMILES, 
mechanism of action, target identifier, and action type. To reduce ambiguity in 
mechanism only descriptions and disambiguate drugs sharing targets, we  constructed an enriched text field, \texttt{text\_rich}, by concatenating  mechanism, target name, action type, and drug preferred name (fetched from the  ChEMBL molecule and target endpoints). Rows with missing SMILES or short  descriptions were removed, yielding $\sim$3k high-quality pairs.

\subsection{Molecular and Text Encoders}
On the molecular side, we experimented with both ChemBERTa \cite{chithrananda2020chemberta} embeddings and ECFP4 fingerprints  (radius $=2$, 2048 bits). On the text side, we used PubMedBERT  \cite{gu2021domain} and a similarity-tuned biomedical encoder  (S-Biomed-RoBERTa-STSB) \cite{deka2021unsupervised}. All encoders 
remained frozen.

\subsection{Thin Contrastive Bridge}
We learn dual linear projection heads (one per modality) that map representations into a shared $d=256$-dimensional space. Training minimizes a symmetric InfoNCE loss with temperature $T=0.07$:
\[
\mathcal{L}
=\tfrac{1}{2}\Big(
\text{CE}\big(\tfrac{B_T B_M^\top}{T}, \text{diag}\big)
+\text{CE}\big(\tfrac{B_M B_T^\top}{T}, \text{diag}\big)\Big),
\]
where $B_T=\text{norm}(W_T Z_T)$ and $B_M=\text{norm}(W_M X_M)$. 
We optimize with AdamW (lr $=10^{-3}$, weight decay $10^{-4}$) for 
100 epochs with batch size 512. To address same-target confusion, 
we incorporate hard-negative weighting and a margin loss.

\subsection{Evaluation Protocols}
We report \textbf{Recall@1}, \textbf{Mean Reciprocal Rank (MRR)}, and 
\textbf{Grouped Recall@1} (within-target discrimination, excluding groups 
with fewer than 3 compounds). We also evaluate under a rigorous  \textbf{scaffold split}, partitioning compounds by Bemis-Murcko scaffold 
into disjoint train/test sets to measure generalization across unseen 
chemical cores. Bootstrap resampling provides 95\% confidence intervals.  

\section{Results}

\subsection{Baseline with Frozen Encoders}
Direct alignment using ChemBERTa for SMILES and PubMedBERT for biomedical text 
proved ineffective. Retrieval accuracy was near random (\texttt{Recall@1 = 0.000--0.001, 
MRR = 0.003}), with similarity tuned S-Biomed-RoBERTa offering no meaningful improvement. 
This confirms that pretrained unimodal encoders alone cannot provide cross-modal alignment.

\subsection{Contrastive Bridge Improves Alignment}
Introducing a lightweight dual projection bridge significantly improved retrieval. 
On the ChEMBL drug mechanism dataset, the bridge achieved 
\textbf{Recall@1 = 0.188} and \textbf{MRR = 0.338}. 
Cosine similarity matrices exhibited a clear diagonal structure after training, 
indicating successful one-to-one molecule text alignment 
(Figure~\ref{fig:bridge_heatmap}).

\subsection{ECFP4 Bridge with Enriched Text}
\begin{figure}[t]
    \centering
    \includegraphics[width=\linewidth]{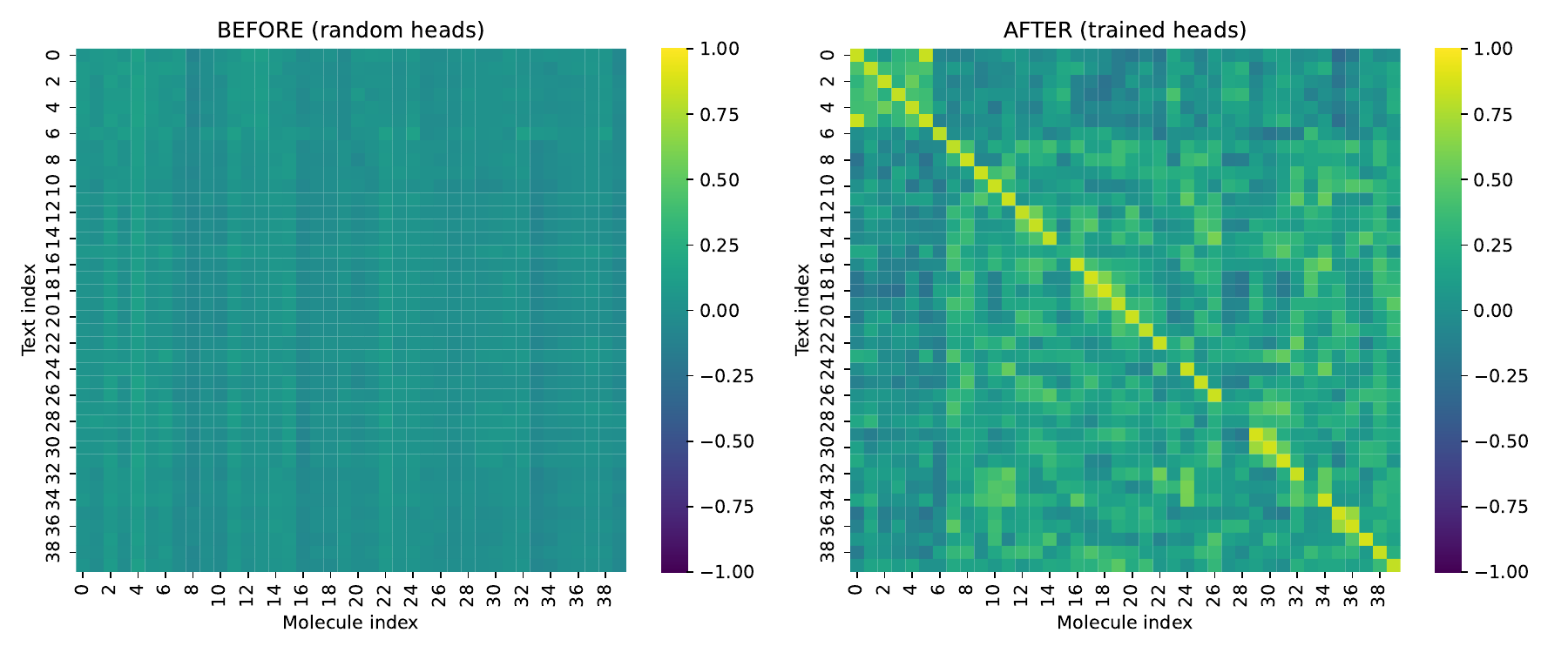}
    \caption{ECFP4 bridge with enriched text (\texttt{text\_rich}). 
    \textbf{Left:} before training. 
    \textbf{Right:} after training, showing clear diagonal alignment.}
    \label{fig:ecfp4_before_after}
\end{figure}

Combining ECFP4 molecular fingerprints with enriched \texttt{text\_rich} descriptions 
substantially boosted alignment. The model achieved 
\textbf{Recall@1 = 0.762} and \textbf{MRR = 0.863}, with strong diagonal structure 
emerging in cosine similarity matrices (Figure~\ref{fig:ecfp4_before_after}). 
Including drug names in \texttt{text\_rich} further improved disambiguation.

\subsection{Generalization under Scaffold Split}
\begin{figure}[t]
    \centering
    \includegraphics[width=\linewidth]{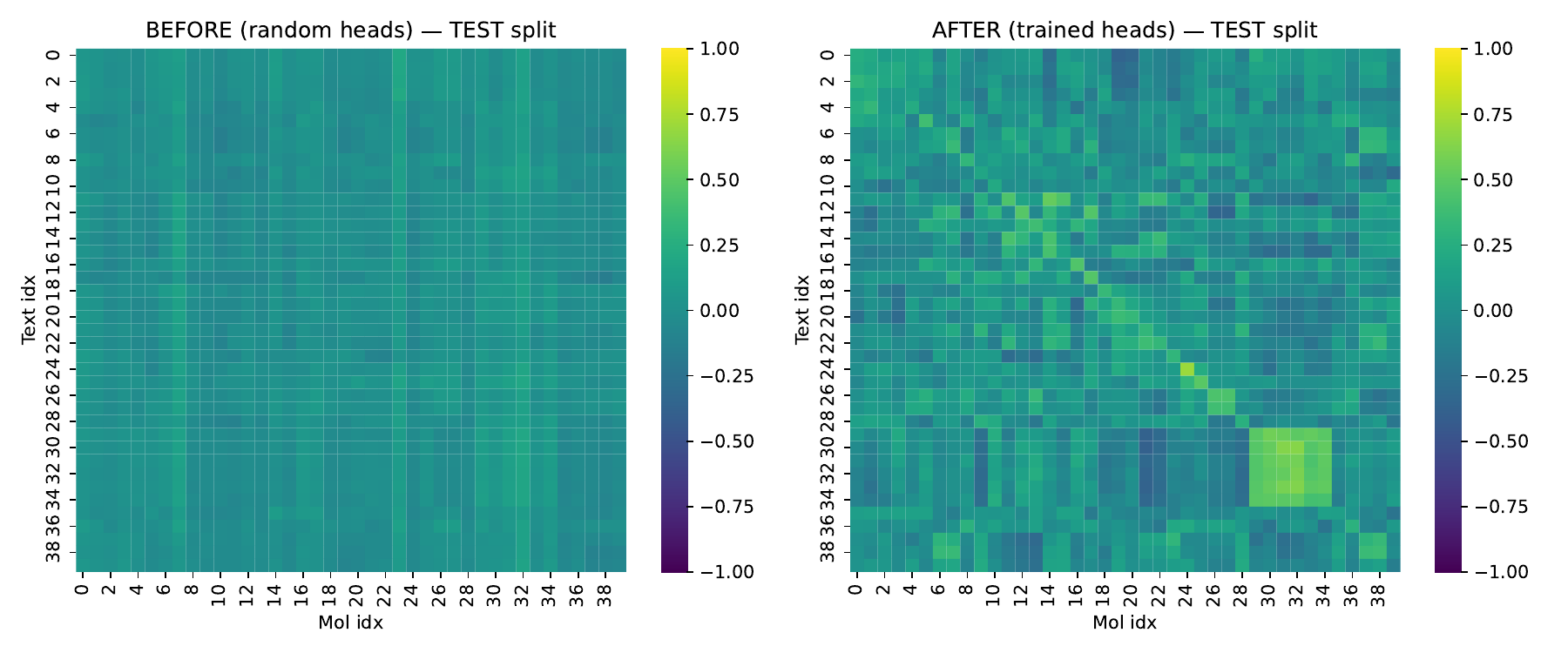}
    \caption{Cosine similarity matrices on the scaffold split test set (first $K{=}40$ pairs).
    \textbf{Left:} random heads. 
    \textbf{Right:} trained bridge with strong diagonal alignment.}
    \label{fig:scaffold_split_heatmap}
\end{figure}

When evaluated under scaffold splits, performance decreased but remained meaningful: 
\textbf{Recall@1 = 0.150}, \textbf{MRR = 0.228}, and 
\textbf{Grouped Recall@1 = 0.317}. 
This more than tripled grouped Recall@1 compared to frozen baselines, 
demonstrating that the bridge generalizes beyond memorized scaffolds 
(Figure~\ref{fig:scaffold_split_heatmap}).

\subsection{Top-$k$ Retrieval and Ablations}
\begin{figure*}[t]
    \centering
    \begin{minipage}[t]{0.48\textwidth}
        \centering
        \includegraphics[width=\linewidth]{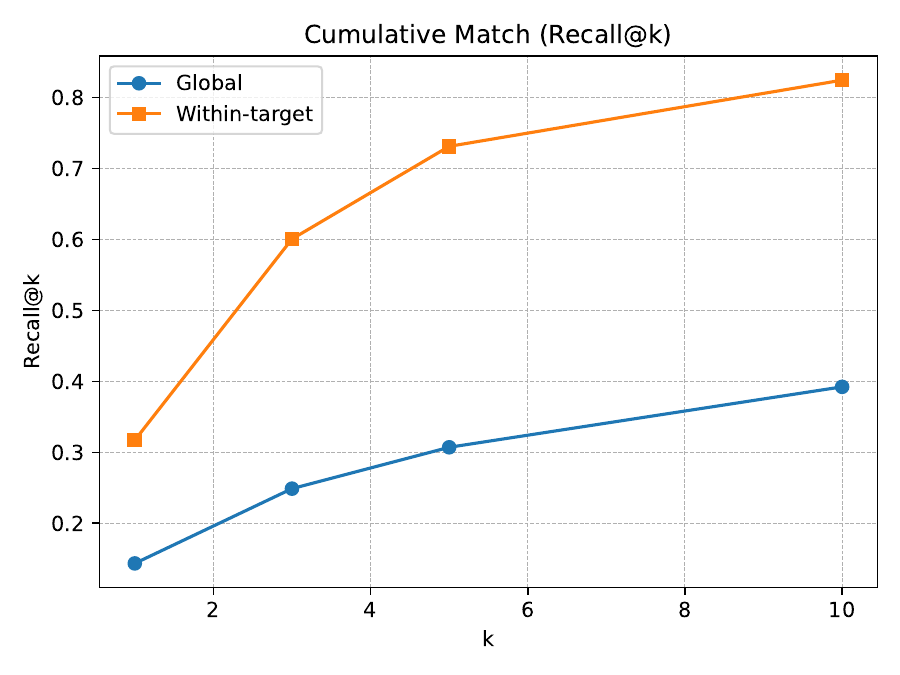}
        \caption{Cumulative Match (Recall@$k$) on the scaffold split test set.
        Global retrieval (blue) improves steadily with $k$, while within-target retrieval
        (orange) climbs steeply, showing most correct matches appear in small sets.}
        \label{fig:cmc_curves}
    \end{minipage}\hfill
    \begin{minipage}[t]{0.48\textwidth}
        \centering
        \includegraphics[width=\linewidth]{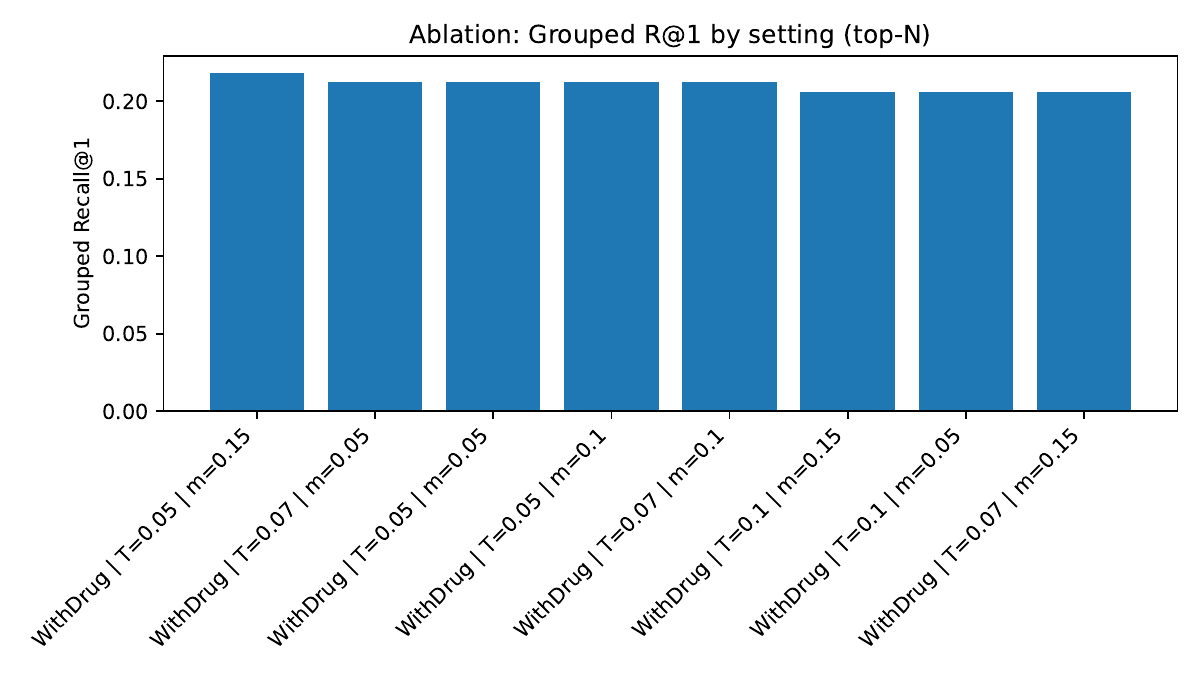}
        \caption{Ablation on grouped Recall@1 across temperature $T$, margin $m$,
        and drug name inclusion in \texttt{text\_rich}. Best: \emph{WithDrug, $T{=}0.05$, $m{=}0.15$}.}
        \label{fig:ablation_bar}
    \end{minipage}
\end{figure*}

Cumulative Match (Recall@$k$) curves show steady improvements with $k$: global Recall@$10 \approx 0.39$, while within-target Recall@$10 > 0.80$,  indicating that the correct molecule is typically found in a small candidate set. Ablations confirmed robustness across temperature $T$ and margin $m$, with best grouped  Recall@1 at \emph{WithDrug, $T{=}0.05$, $m{=}0.15$}  (Figure~\ref{fig:cmc_curves}, Figure~\ref{fig:ablation_bar}).

\subsection{Summary of Results}
Table~\ref{tab:results} summarizes performance across settings.

\begin{table}[H]
\centering
\caption{Retrieval performance on ChEMBL drug--text alignment.}
\label{tab:results}
\begin{tabular}{lccc}
\toprule
\textbf{Setting} & \textbf{Recall@1} & \textbf{MRR} & \textbf{Grouped R@1} \\
\midrule
Frozen encoders (ChemBERTa + PubMedBERT) & 0.001 & 0.003 & -- \\
Frozen encoders (ChemBERTa + S-Biomed)   & 0.000 & 0.003 & -- \\
Contrastive bridge (ChemBERTa + PubMedBERT) & 0.188 & 0.338 & 0.098 \\
ECFP4 + \texttt{text\_rich} bridge & \textbf{0.762} & \textbf{0.863} & -- \\
Scaffold split (ECFP4 + \texttt{text\_rich}) & 0.150 & 0.228 & 0.317 \\
\bottomrule
\end{tabular}
\end{table}

\subsection{Future Directions}
Our results suggest that cross modal alignment can accelerate drug discovery. 
By enabling efficient retrieval of molecule mechanism pairs, the framework 
supports rapid in silico screening and provides interpretable links between 
chemical structure and biological function. Moreover, scaffold split generalization 
indicates potential for discovering compounds with novel scaffolds. Coupling this 
approach with generative molecular models could further enable \emph{de novo} 
design of candidate drugs.

\bibliographystyle{plainnat} 
\bibliography{references}    


\clearpage
\appendix
\section{Appendix}

\begin{figure}[H]
  \centering
  \includegraphics[width=0.55\linewidth]{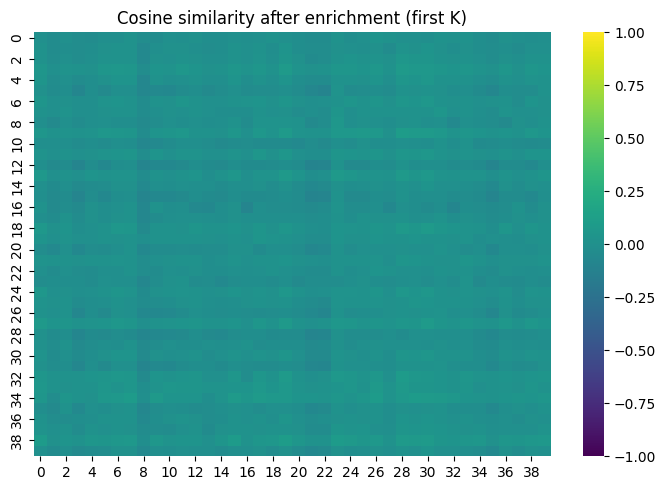}
  \caption{Cosine similarity heatmap between the first $K=40$ drug--text pairs after enrichment. The uniform pattern indicates that frozen encoders alone do not yield meaningful alignment.}
  \label{fig:cosine_heatmap}
\end{figure}

\begin{figure}[H]
  \centering
  \includegraphics[width=0.95\linewidth]{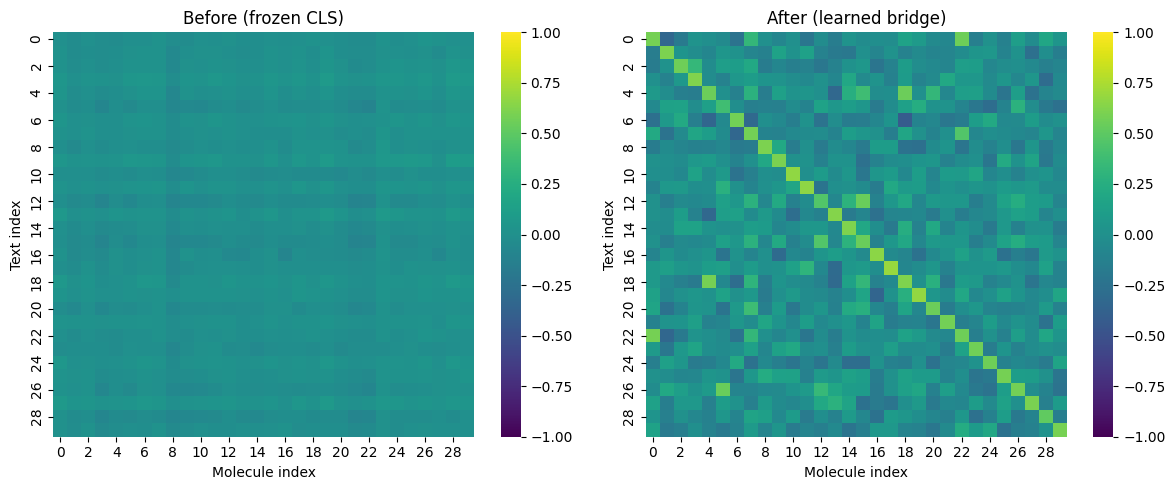}
  \caption{Cosine similarity comparison between the first $K=30$ drug--text pairs before and after training. (\textbf{Left}) Frozen CLS embeddings show uniform similarity with no meaningful alignment. (\textbf{Right}) After training the learned bridge, a strong diagonal emerges, indicating accurate molecule--text alignment.}
  \label{fig:bridge_heatmap}
\end{figure}

\end{document}